\pdfoutput=1
\documentclass[10pt, a4paper]{article}
\usepackage{lt4all}
\usepackage{booktabs}
\usepackage{url}
\usepackage[table]{xcolor}
\definecolor{light-gray}{gray}{0.95}
\usepackage{microtype}
\usepackage{xspace}
\usepackage{xstring}
\usepackage{refcount}
\usepackage{tabulary}

\title{Google Crowdsourced Speech Corpora and Related Open-Source
  Resources for Low-Resource Languages and Dialects: An Overview}

\name{%
  Alena Butryna,
  Shan-Hui Cathy Chu,
  I{\c{s}}{\i}n Demir{\c{s}}ahin,
  Alexander Gutkin,
  Linne Ha$^{\dagger}$\thanks{\mdseries\small$^\dagger$The author contributed to this work while at Google.},
  \\\bfseries\large%
  Fei He,
  Martin Jansche$^{\dagger}$,
  Cibu Johny,
  Anna Katanova,
  Oddur Kjartansson,
  \\\bfseries\large%
  Chenfang Li$^{\dagger}$,
  Tatiana Merkulova,
  Yin May Oo$^{\dagger}$,
  Knot Pipatsrisawat,
  Clara Rivera,
  \\\bfseries\large%
  Supheakmungkol Sarin,
  Pasindu de Silva,
  Keshan Sodimana$^{\dagger}$,
  \\\bfseries\large%
  Richard Sproat,
  Theeraphol Wattanavekin,
  Jaka Aris Eko Wibawa$^{\dagger}$%
}

\address{
  Google Research, Singapore, United States and United Kingdom \\
  \{alenab,oddur,rivera,mungkol\}@google.com}

\abstract{
  This paper presents an overview of a program designed to address the
  growing need for developing freely available speech resources for
  under-represented languages. At present we have released 38 datasets
  for building text-to-speech and automatic speech recognition
  applications for languages and dialects of South and Southeast Asia,
  Africa, Europe and South America. The paper describes the
  methodology used for developing such corpora and presents some of
  our findings that could benefit under-represented language
  communities.}

\begin{document}

\maketitleabstract


\section{Introduction}
Historically speech and language technology research has focused on a
few major Indo-European languages, along with Mandarin and
Japanese. The past decade, however, has seen an increased focus by the
speech and language research community and technological companies on
addressing the plight of low resource and especially the endangered
languages. According to various sources, such as~\newcite{ethnologue},
roughly 40\% of 5,000 to 7,000 languages spoken today are classified
as endangered. The shift of focus is partly due to the awareness of
the importance of preserving and documenting the languages which are
at risk of losing its last native speakers due to the shift to other
dominant languages or disappearance of the communities altogether. In
addition to the critically endangered languages, there are hundreds of
languages with large native speaker populations which are classified
as low-resource (sometimes low-density) due to the lack of linguistic
resources necessary for advancing research and technological
innovation. Furthermore, the Internet is creating a large and growing
divide between languages that are represented in technology and those
that are not: it is estimated that only 5\% of the world’s languages
are accessible on the Internet.

In this paper we offer a brief overview of a linguistic program
which aims to provide free speech resources in regions with fast
growing Internet communities but few publicly available linguistic
resources. So far the efforts have mostly focused on statutory
national or provincial languages, with the overall goal of
investigating the methods to scale our approach to many smaller
regional languages in the locales of interest. One of the primary
goals of the program is to develop an accessible and replicable
methodology that any local community of technologists can use with
available open-source solutions to build custom applications utilizing
our released resources or to construct their own resources for a new
language. At the same time, it is important to make sure that the
quality of the resulting solutions built using this methodology are on
par with the systems for better-resourced languages.

Another component of this program deals with the construction of
corpora for low-resource dialects of well-resourced languages. More
often than not, the assumption exists that a local community is
adequately served by providing the speech technology built using the
dominant dialect, despite the dialects significantly diverging to the
point of being mutually unintelligible (e.g., High German vs. Swiss
German). An interesting part of this process is investigating the
optimal strategies for constructing local dialect-specific resources
that build upon the existing well-resourced language resources in a
way that adequately serves the local community.

The program focuses on developing the resources for two types of
applications: automatic speech recognition (ASR) and text-to-speech
(TTS), both of which are crucial components of modern technological
ecosystems for any given language. These applications have different
resource requirements: the modern ASR systems typically require more
data from as many speakers as possible, while the TTS systems ideally
need higher-quality recordings from fewer speakers but with well
articulated speech. Additional resources, such as text normalization
grammars for converting non-standard word tokens to natural language
words and models of phonology are often needed as well. These are
highly language-specific and require considerable linguistic expertise
to develop.


\section{Program Overview}

\subsection{Selection of Languages and Dialects}
The languages and dialects selected for the program are broadly based
on two selection criteria. The first goal is to increase the
availability of open-source speech resources in the regions which were
identified as important (in terms of number of speakers, Internet
penetration and cultural significance for the region) and yet
considered low-resource. The initiatives for constructing linguistic
resources for such languages, whether from local or foreign
governments or big technological companies, can effectively tip the
balance and cause the language to become well-resourced, as
happened with Modern Standard Arabic in the course of the last twenty
years. The second criterion involves selection of languages from
diverse language families so that the generality of resource
collection paradigm can be optimized based on the exposure to
different linguistic and operational requirements and the findings
shared with the community. The TTS and ASR corpora collected so far
consist of over 1,500 hours of speech and are freely available online
hosted by~\newcite{openslr} under unencumbered license~\cite{license}.

\paragraph{South Asia}
The South Asian languages selected for the program include the
languages from Indo-Aryan and Dravidian language families. The
Indo-Aryan languages selected so far include two dialects of Bengali
(India and Bangladesh), Gujarati (India), Marathi (India), Nepali
(Nepal) and Sinhala (Sri Lanka). The set of Dravidian languages
included by the program include Kannada (India), Malayalam (India),
Tamil (India) and Telugu (India). According to various estimates,
these languages have a combined population of about 706 million native
and second-language speakers. From a research standpoint, these
languages are very interesting to work with: they exhibit considerable
variation within each language family, but at the same time also have
considerable similarities across both language families.

\paragraph{Southeast Asia}
The set of Southeast Asian languages we selected includes Burmese
(Myanmar) from Sino-Tibetan language family, Khmer (Cambodia) from
Austroasiatic language family, and Javanese (Indonesia) and Sundanese
(Indonesia) from Malayo-Polynesian language family. These languages
are natively spoken by about 178 million people across the region. The
language families in this set are very diverse and yet exhibit
considerable influence from their neighbors from other language
families in both South and Southeast Asia.

\paragraph{Africa}
Four out of eight official languages of South Africa were selected:
Sesotho, Setswana, Xhosa (from a Bantu language family) and Afrikaans
(Indo-European). These languages have a combined speaker population of
native and second-language speakers of about 62.5 million. In addition
we selected Nigerian English as one of the largest and yet low-resource
dialects of English on the continent.

\paragraph{Europe and South America}
To increase the coverage of Indo-European language family among the
selected languages, a set of three regional languages of Spain with
combined population of about 14 million speakers were selected:
Galician and Catalan, both of which belong to Ibero-Romance group, and
Basque, which is a language isolate. As part of our work towards
improving the availability of open-source speech resources for
low-resource dialects and regional accents of the better served
languages, we also collected speech corpora for six Latin American
Spanish dialects (Argentinian, Colombian, Chilean, Peruvian, Puerto
Rican and Venezuelan) and various dialects and accents of Irish and
British English (Welsh English, Southern English, Midlands English,
Northern English and Scottish English).

\subsection{Methodology}

\paragraph{Local Community and University Outreach}
It goes without saying that any collaborative corpora collection is
greatly helped by the enthusiastic community of native
speakers. Throughout the program we tried to enlist the help from
local universities, technology and language enthusiasts wherever
possible. This approach is illustrated by the data collection process
for Javanese and Sundanese for which the collaboration with two local
universities was established. For Javanese we worked with the Faculty
of Computer Science at Universitas Gadjah Mada (UGM) in Yogyakarta,
while for Sundanese a collaboration with the Faculty of Language and
Literature at Universitas Pendidikan Indonesia (UPI) in Bandung was
established. The universities assisted us with finding volunteers to
help manage the data collection, as well as with the adequate
recording environments. The university staff put us in contact with
the student organizations which helped to disseminate the information
about corpus collection and call for volunteers. A portion of the
recordings was done at the student-run annual Computer Science
exhibition event organized by students from the Faculty of Computer
Science at Universitas Indonesia~\cite{wibawa18building}. A similar
approach was followed in South Africa where we established the
collaboration with Multilingual Speech Technologies group from the
North-West University to assemble the speech corpora for four South
African languages~\cite{niekerk17rapid}. In other countries, such as
Bangladesh, Cambodia, Myanmar, Nepal and Sri Lanka, we followed the
same blueprint involving the participants from local universities in
the processes of data collection and
curation~\cite{kjartansson2018}. In Latin America, we worked with
local Google Developer group representatives.

\paragraph{Software and Hardware Equipment}
For ASR data collections, we required recording applications capable
of running on low-end smartphone devices. While initially we relied on
a proprietary application, we later teamed up with University of
Reykjavik in Iceland and migrated our ASR collections to use their
open-source software~\cite{petursson2016}. Since TTS corpora requires
higher speech quality, we went through several careful iterations to
settle on a hardware combination that was lightweight and portable
while providing the best possible quality for our
purposes. Additionally, we made sure that the equipment in question
was affordable to local communities. One of the configurations that
was found to work well for us and that we recommend to others includes
an ASUS Zenbook UX305CA fanless laptop, Neumann KM 184 microphone, a
Blue Icicle XLR-USB A/D converter and a portable acoustic booth. The
overall cost of this configuration, especially when reused for
multiple data collections, is well below the cost of renting a
professional recording studio.

\paragraph{Development of Recording Materials}
Open sourcing low-resource language speech corpora was of high
priority since the inception of the program. Therefore, during the
recording script development we made sure we use publicly available
sources. For both ASR and TTS corpora Wikipedia text was used in the
form of short sentences extracted at random (when available in the
language of interest, otherwise crowdsourced translations thereof). In
addition, for TTS recording scripts further combination of three types
of materials was prepared: (1) Handcrafted text to ensure a broad
phonetic coverage of the language, filling in any gaps from Wikipedia,
(2) template sentences including common named entities and numeric
expressions in each language. These were obtained in collaboration
with communities and partners and include celebrity names,
geographical names, telephone numbers, time expressions and so on, (3)
domain specific real-world sentences that could be used in product
applications, usually covering navigation, sports or weather.

\paragraph{Recording and Quality Control Procedures}
The ASR corpora are collected using standard consumer smartphones. No
specialized or additional hardware is used for the data
collection. The collections are overseen by the volunteer field
workers who are trained to guide the volunteer speakers. During the
recording the audio is first saved to local storage on the device and
then uploaded to a server once a connection to the Internet is
established. This feature of the recording software is important
because limited Internet connectivity potentially poses serious
operational problems to the field workers, especially in remote areas.

Recording of TTS corpora poses different challenges because the goal
is to collect high-quality and well-articulated speech samples in a
portable recording studio. A short introduction is given to
participants so that they can confidently operate the software during
the recording session, alongside with the guidelines on the relative
position of the volunteer speaker to the microphone to ensure
consistency between different recording sessions.

Since none of the speakers recorded for TTS are professional voice
talents, their recordings often contain problematic artifacts such as
unexpected pauses, spurious sounds (like coughing) and breathy
speech. All recordings go through a quality control process performed
by the trained native speakers to ensure that each utterance matches
the corresponding transcript, has consistent volume, is noise-free and
consists of fluent speech without unnatural pauses.

\paragraph{Other Types of Linguistic Resources}
Building competitive ASR and TTS applications for low-resource
languages typically requires the development of further linguistic
resources in addition to corpora and our program takes this
requirement into account. The necessary components required for
building robust speech ecosystem for any given language typically
include carefully designed phonological representation upon which the
pronunciation lexicons can be based, the algorithms for generating
pronunciations for words missing from the lexicon and a system for
converting between non-standard word (NSW) tokens, such as numbers,
and the corresponding natural language words. The development of these
components typically requires native knowledge of the language and
considerable linguistic expertise. Therefore we are making sure that
any additional linguistic artifacts developed by the program are well
documented and freely accessible~\cite{glr}. Examples of such
artifacts include phonological representation for Lao, pronunciation
guidelines for Burmese and text normalization grammars for languages
of South and Southeast Asia~\cite{sodimana18text}.

\subsection{Emerging Lessons}

\paragraph{Operational Lessons}
Throughout the program's lifetime we continuously discovered the
positive impact the collaboration with local communities had on our
data collection projects. This is partly due to local technologists,
academics and open-source enthusiasts who understand well how the
availability of technology in their local language can positively
impact the life of a local community and often enthusiastically
endorse initiatives such as ours. In our particular case, setting up
the crowd-sourcing mechanisms locally would have not been possible
without their support.

Furthermore, such collaborations often resulted in important
contributions to other Google programs. For example, simultaneously
with collecting the speech corpora in Bangladesh, Cambodia, Nepal,
Myanmar and Sri Lanka, we hosted a series of media workshops,
train-the-trainer sessions and translate-a-thons with universities and
community groups, that aimed to educate people about how they can use
the new Google Translate Community tool~\cite{gtranslate} to improve
the accuracy, understanding and representation of their language on
the web. We also actively participated and contributed to various
local conferences primarily focused around technology and the web.

Finally, we discovered that using moderately priced recording
equipment was enough to collect the corpora of sufficient quality to
suit the needs of local community and, at the same time, also be used
in real Google products.

\paragraph{Research and Development Findings}
One of the first findings of this program is that the crowd-sourced
TTS corpora works adequately in a single multi-speaker
model~\cite{gutkin2016}. While the quality of the resulting model was
somewhat below the quality of state-of-the-art commercial systems, at
later stages of the program we discovered that the quality of such
models can be significantly improved by combining multi-speaker
crowdsourced corpora from multiple languages. This finding capitalizes
on the notion of “linguistic area” by an eminent American
linguist~\newcite{emeneau56}, where he defines it (p. 16, fn. 28) as
``an area which includes languages belonging to more than one language
family but showing traits in common which are found to belong to the
other members of (at least) one of the families''. Based on this
observation we successfully built multilingual system based on the
combined corpus of South Asian Indo-Aryan and Dravidian
languages~\cite{demirsahin18unified} and Malayo-Polynesian
multilingual system combining our Javanese and Sundanese corpora with
a proprietary corpus of
Indonesian~\cite{wibawa18building}. Furthermore, we found that our
South Asian multilingual model was good enough for synthesizing the
languages for which we had no training data, such as Odia and Punjabi.

Another lesson that emerged from applying the collected speech corpora
in practical applications is the importance of freely available
typological resources, such as PHOIBLE~\cite{phoible}. During work on
low-resource language technology, more often than not, the required
linguistic (and in particular, phonological) expertise is hard to
find, even among the native speakers. The availability of typological
resources as a reference have significantly boosted our research and
development efforts.

\paragraph{Important Challenges}
While working on this program, we have identified several important
areas which need to be addressed in order to scale this work to many
more languages and dialects which currently possess even fewer
linguistic resources than the languages we have dealt with so far.

When it comes to developing speech corpora and applications, more
often than not, the ``one size fits all'' approach does not work
because different language families present very different
challenges. Once the speech corpora are collected, the types of
language-specific challenges that may block application development
include the lack of large amounts of labeled training data for
training word segmentation algorithms (e.g., Burmese, Khmer and Lao),
lack of morphosyntactic taggers for smaller Slavic languages
(e.g. Rusyn) required for proper function of text normalization and so
on. Streamlining this process is highly non-trivial because currently
no universal recipe exists.

Moreover, as we mentioned previously, development of linguistic
resources requires considerable linguistic expertise, which is often
hard to find. The deep learning end-to-end approaches, which have
recently gained popularity~\cite{toshniwal2018}, offer potential
workaround. Such systems can be adapted to smaller languages using
transfer learning techniques~\cite{chen2019}. At the same time, such
systems are notoriously data hungry and further techniques utilizing
more data, including lower-quality data found
online~\cite{cooper2019}, may be required. Furthermore, some form of
linguistic knowledge is desirable in such systems due to occasional
unpredictable errors they are prone to~\cite{zhang2019}.


\section{Concluding Remarks}
We have presented an overview of the program that helped collect and
release 38 datasets for building TTS and ASR applications for
languages and dialects of South and Southeast Asia, Africa, Europe and
South America. The corpora collected so far consist of over 1,500
hours of speech and are freely available online. Partnering with local
universities and communities in the region was crucial to the success
of the program as it connected us with a lot of enthusiastic local
contributors which in its turn resulted in collecting high quality
data. We do hope that the described methodology and the released
datasets will be utilized by the local communities to develop custom
applications or to collect new datasets going forward.

There are still many endangered and low resource languages that we
want to focus on in our program. Even though the program already
allows to collect data for language resources development efficiently
from an operational perspective, there are still challenges that need
to be addressed at the development of linguistic resources stage so
that the work can continue at scale. As of now, the program
established a good foundation, however, there is still work to be
done. We hope that these efforts will facilitate future research by
the broader scientific community and will encourage others to apply
our program methodologies and findings to benefit under-represented
language research.


\section{Bibliographical References}
\bibliographystyle{lt4all}
\bibliography{main}

\end{document}